\documentclass[10pt,twocolumn,letterpaper]{article}

\usepackage{iccv}
\usepackage{times}
\usepackage{epsfig}
\usepackage{graphicx}
\usepackage{amsmath}
\usepackage{amssymb}
\usepackage{booktabs}
\usepackage{multirow}
\usepackage{subfigure}
\usepackage{caption}
\usepackage[usenames,dvipsnames]{color}
\usepackage{colortbl}
\usepackage{wrapfig}
\definecolor{mygray}{gray}{.9}
\usepackage{makecell}
 
\usepackage[accsupp]{axessibility} 

\usepackage[breaklinks=true,bookmarks=false]{hyperref}

\iccvfinalcopy 

\def\iccvPaperID{****} 

\ificcvfinal\fi

\begin{document}

\title{Beyond One-to-One: Rethinking the Referring Image Segmentation}

\author{Yutao Hu$^{1}$\footnotemark[1], \quad Qixiong Wang$^{2}$\footnotemark[1], \quad Wenqi Shao$^{2}$, \quad Enze Xie$^{3}$,\\ Zhenguo Li$^{3}$, \quad Jungong Han$^{4}$, \quad Ping Luo$^{1,2}$\footnotemark[2]\\
$^{1}$The University of Hong Kong \quad $^{2}$Shanghai AI Laboratory \\ $^{3}$Huawei Noah’s Ark Lab \quad $^{4}$The University of Sheffield\\
\and
}

\maketitle
\ificcvfinal\thispagestyle{empty}\fi

\footnotetext[1]{Equal contribution.}
\footnotetext[2]{Corresponding author.}

\begin{abstract}
   Referring image segmentation aims to segment the target object referred by a natural language expression. However, previous methods rely on the strong assumption that one sentence must describe one target in the image, which is often not the case in real-world applications. As a result, such methods fail when the expressions refer to either no objects or multiple objects. In this paper, we address this issue from two perspectives. First, we propose a Dual Multi-Modal Interaction (DMMI) Network, which contains two decoder branches and enables information flow in two directions. In the text-to-image decoder, text embedding is utilized to query the visual feature and localize the corresponding target. Meanwhile, the image-to-text decoder is implemented to reconstruct the erased entity-phrase conditioned on the visual feature. In this way, visual features are encouraged to contain the critical semantic information about target entity, which supports the accurate segmentation in the text-to-image decoder in turn. Secondly, we collect a new challenging but realistic dataset called Ref-ZOM, which includes image-text pairs under different settings. Extensive experiments demonstrate our method achieves state-of-the-art performance on different datasets, and the Ref-ZOM-trained model performs well on various types of text inputs. Codes and datasets are available at \url{https://github.com/toggle1995/RIS-DMMI}.

\end{abstract}

\section{Introduction}

Referring image segmentation aims to segment the target object described by a given natural language expression. Compared to the traditional semantic segmentation task \cite{long2015fully, ronneberger2015u, chen2017rethinking}, referring image segmentation is no longer restricted by the predefined classes and could segment specific individuals selectively according to the description of text, which has large potential value for various applications such as human-robot interaction \cite{wang2019reinforced} and image editing \cite{chen2018language}. Despite the recent progress, there are still several important challenges that need to be addressed in order to make this technology more applicable in real-world scenarios.

\begin{figure}[tbp] 
\centering 
\includegraphics[width=1\linewidth]{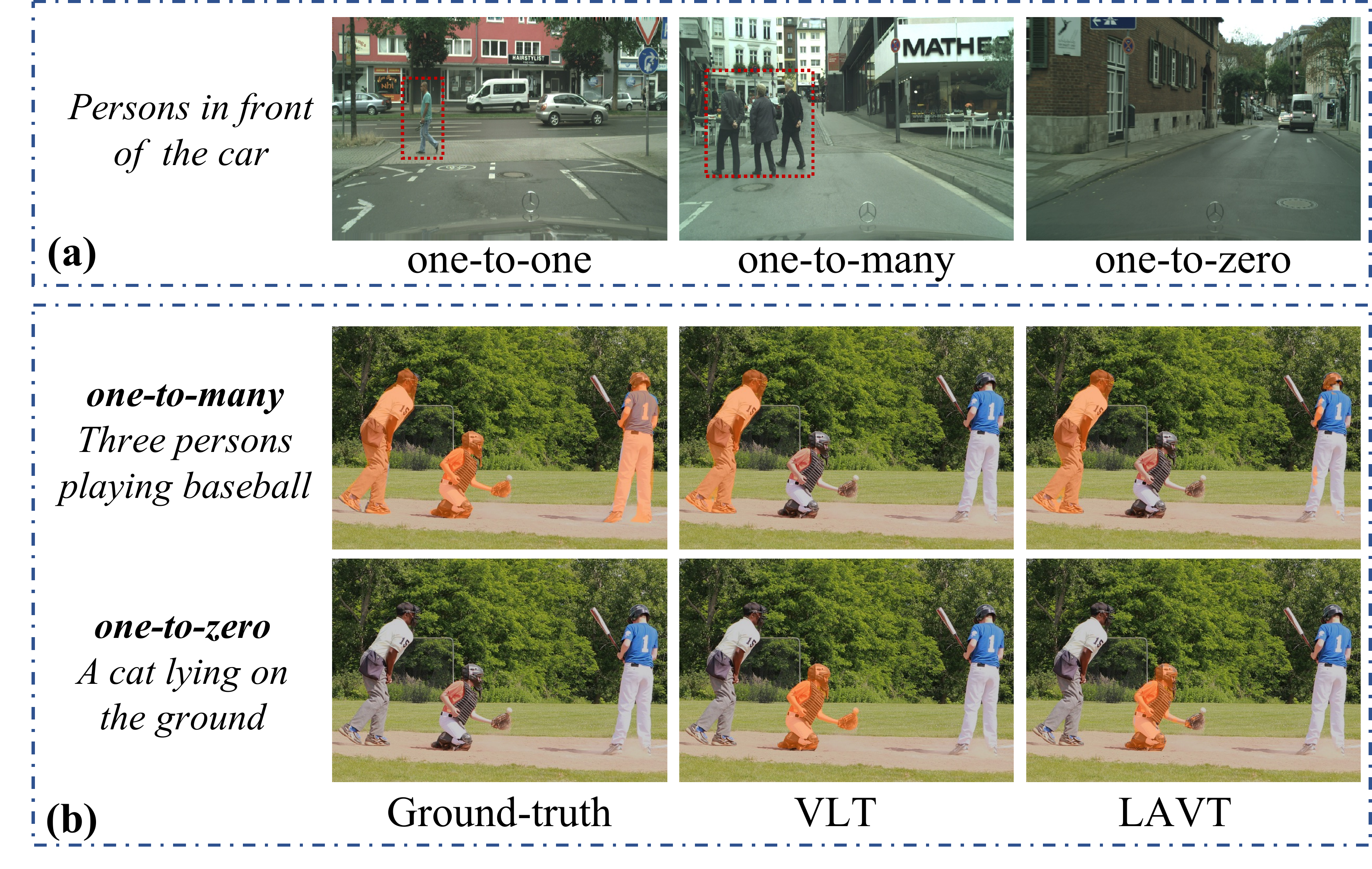} 
\put (-14,10) {\makebox(0,0){\footnotesize\cite{yang2022lavt}}}
\put (-78,10) {\makebox(0,0){\footnotesize\cite{ding2021vision}}}
\caption{(a) Taking the autonomous driving as an example, text expression may refer to varying number of targets, depending on the specific real-world scenario. (b) When the sentences refer to multiple or no targets, existing methods cannot realize accurate segmentation.} 
\label{fig:moti} 
\vspace{-3mm}
\end{figure}

In referring image segmentation, most previous methods only concentrate on the one-to-one setting, where each sentence only indicates one target in the image. However, as shown in Fig.~\ref{fig:moti}(a), one-to-many and one-to-zero settings, where the sentence indicates many or no targets in the image, respectively, are also common and critical in the real-world applications. Unfortunately, previous methods tend to struggle when confronting one-to-many and one-to-zero samples. As illustrated in Fig.~\ref{fig:moti}(b), the recent SOTA method, LAVT \cite{yang2022lavt}, only localizes one person in the image when given the description ``Three persons playing baseball". As for one-to-zero input, previous methods still segment one target even if it is completely irrelevant to the given text. Therefore, it is imperative to enable the model to adapt to various types of text inputs.

We attribute this problem to two main factors. First, although existing methods design various ingenious modules to align multi-modal features, most of them only supervise the pixel matching of the segmentation map, which cannot ensure the significant semantic clues from the text are fully incorporated into the visual stream. As a result, visual features lack the comprehensive understanding of the entity being referred to in the expression, which limits the capacity when the model confronts various types of text inputs. Second, all popular datasets \cite{kazemzadeh2014referitgame, nagaraja2016modeling, mao2016generation} for referring image segmentation are established under the one-to-one assumption. In the training, the model is enforced to localize one entity that is most related to the text. As a result, the model trained on these datasets is prone to overfitting and only remembers to segment the object with the largest response, which leads to the failure when segmenting one-to-many and one-to-zero samples. 

To address the aforementioned issues, this paper proposes a Dual Multi-Modal Interaction Network (DMMI) to achieve robust segmentation when given various types of text expressions, and establishes a new comprehensive dataset Ref-ZOM (\textbf{Z}ero/\textbf{O}ne/\textbf{M}any). In the DMMI network, we address the referring segmentation task in a dual manner, which not only incorporates the text information into visual features but also enables the information flow from visual stream to the linguistic one. As illustrated in Fig.~\ref{fig:framework}, the whole framework contains two decoder branches. On the one hand, in the text-to-image decoder, linguistic information is involved into the visual features to segment the corresponding target. On the other hand, we randomly erase the entity-phrase in the original sentence and extract the incomplete linguistic feature. Then, in the image-to-text decoder, given the incomplete text embedding, we utilize the Context Clue Recovery (CCR) module to reconstruct the missing information conditioned on the visual features. Meanwhile, multi-modal contrastive learning is also deployed to assist the reconstruction. By doing so, the visual feature is encouraged to fully incorporate the semantic clues about target entity, which promotes the multi-modal feature interaction and leads to more accurate segmentation maps. Additionally, to facilitate the two decoder parts, we design a Multi-scale Bi-direction Attention (MBA) module to align the multi-modal information in the encoder. Beyond the interaction between single-pixel and single-word \cite{yang2022lavt}, the MBA module enables the multi-modal interaction in the local region with various sizes, leading to a more comprehensive understanding of multi-modal features.

In the Ref-ZOM, we establish a comprehensive and challenging dataset to promote the referring image segmentation when given various types of text inputs. On the one hand, compared to the existing widely-used datasets \cite{kazemzadeh2014referitgame, nagaraja2016modeling, mao2016generation}, the text expressions are more complex in Ref-ZOM. It is not limited to the one-to-one assumption, and instead, the expression can refer to multiple or no targets within the image. Additionally, the language style in our Ref-ZOM is much more flowery than the short phrases found in \cite{kazemzadeh2014referitgame}. On the other hand, Ref-ZOM also surpasses most mainstream datasets in terms of size, containing 55078 images and 74942 annotated objects.

We conduct extensive experiments on three popular datasets \cite{kazemzadeh2014referitgame, nagaraja2016modeling, mao2016generation} and our DMMI achieves state-of-the-art results. Meanwhile, we reproduce some representative methods on our newly established Ref-ZOM dataset, where DMMI network consistently outperforms existing methods and exhibits strong ability in handling one-to-zero and one-to-many text inputs. Moreover, the Ref-ZOM-trained network performs remarkable generalization capacity when being transferred to different datasets without fine-tuning, highlighting its potential for real-world applications.

The main contributions of this paper are summarized as follows:

\begin{itemize}	
\item We find the deficiency of referring image segmentation when meeting the one-to-many and one-to-zero text inputs, which strongly limits the application value in real-world scenarios.
	
\item We propose a Dual Multi-Modal Interaction (DMMI) Network to enable the information flow in two directions. Besides the generation of segmentation map, DMMI utilizes the image-to-text decoder to reconstruct the erased entity-phrase, which facilitates the comprehensive understanding of the text expression.
	
\item We collect a new challenging dataset, termed as Ref-ZOM, in which the text inputs are not limited to the one-to-one setting. The proposed dataset provides a new perspective and benchmark for future research.

\item Extensive experimental results show the proposed DMMI network achieves new state-of-the-art results on three popular benchmarks, and exhibits superior capacity in handling various types of text inputs on the newly collected Ref-ZOM.

\end{itemize}

\section{Related Work}
\subsection{Referring Image Segmentation}

Referring image segmentation is first introduced by \cite{hu2016segmentation}. Early approaches \cite{hu2016segmentation, li2018referring, liu2017recurrent, margffoy2018dynamic} generally employ Convolutional Neural Networks (CNNs) \cite{chen2017rethinking, redmon2018yolov3, he2016deep} and Recurrent Neural Networks (RNNs) \cite{hochreiter1997long, huang2015bidirectional} to extract relevant visual and linguistic features. After feature extraction, the concatenation-convolution operation is employed to fuse multi-modal features. However, it fails to exploit the inherent interaction between image and text. To overcome this shortcoming, some approaches \cite{huang2020referring, hui2020linguistic, yang2021bottom} establish relation-aware reasoning based on the multi-modal graph. 

\begin{figure*}[tbp] 
\centering 
\includegraphics[width=1\linewidth]{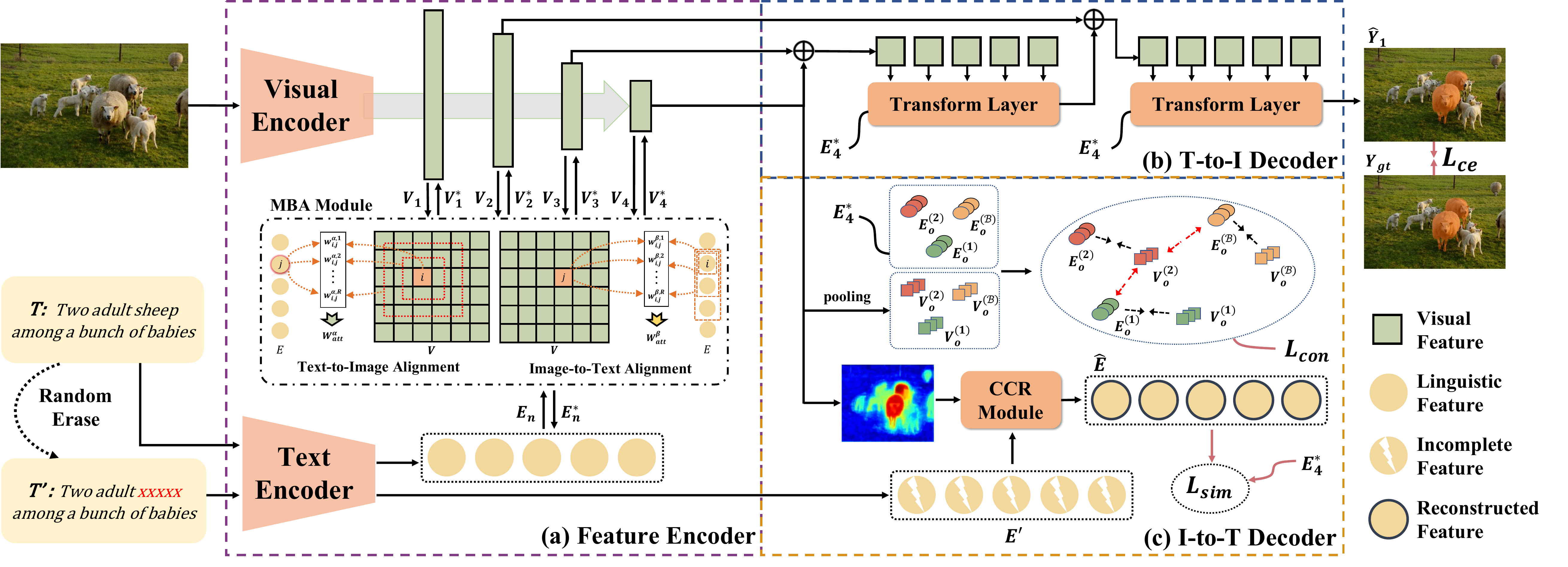}
\vspace{-7mm}
\caption{The whole framework of the proposed Dual Multi-Modal Interaction (DMMI) Network. \textbf{(a)} The feature encoder, in which the visual encoder and text encoder are utilized to extract visual and linguistic feature, respectively. Meanwhile, MBA module is employed to perform multi-modal feature interaction. Notably, $w_{i,j}$ denotes the $j$-th point in the $i$-th row of attention weight. \textbf{(b)} Text-to-image decoder, in which the text embedding is utilized to query the visual feature to generate the prediction map. \textbf{(c)} Image-to-text decoder, in which CCR module is utilized to reconstruct the erased linguistic information condition on the visual feature. $\mathcal{L}_{con}$ and $\mathcal{L}_{sim}$ are implemented to assist the reconstruction.}
\vspace{-5mm}
\label{fig:framework} 
\end{figure*}

Recently, due to the breakthrough of Transformer in computer vision community \cite{dosovitskiy2020image, xie2021segformer, han2022survey, carion2020end}, Transformer-based backbones have become dominant in referring image segmentation for both visual and linguistic feature extraction \cite{liu2021swin, devlin2018bert, jing2021locate, kamath2021mdetr}. Meanwhile, the self-attention mechanism \cite{vaswani2017attention} in the Transformer also inspired numerous studies that employ cross-attention blocks for better cross-model alignment. For instance, VLT \cite{ding2021vision} utilizes the cross-attention module to generate the query vectors by comprehensively understanding the multi-model features, which are then used to query the given image through the Transformer decoder. LAVT \cite{yang2022lavt} finds that early fusion of multi-modal features via cross-attention module brings better cross-modal alignments. Moreover, CRIS \cite{wang2022cris} utilizes the Transformer block to transfer the strong ability of image-text alignment from the pre-trained CLIP model \cite{radford2021learning}.

However, most previous methods only supervise the visual prediction and cannot ensure the semantic clues in the text expressions have been incorporated into the visual features. As a result, these methods tend to struggle when handling the text expressions that refer to either no objects or multiple objects. In this work, we establish a dual network and emphasize that the information flow from image to text is beneficial for comprehensive understanding of text expression. Furthermore, we collect a new dataset called Ref-ZOM, which contains various types of text inputs and compensates for the limitations of existing benchmarks.

\subsection{Visual-Language Understanding}
Video-Language understanding has received rapidly growing attention in recent years and plays an important role in various tasks such as video-retrieval \cite{gabeur2020multi}, image-text matching \cite{li2019visual} and visual question answering \cite{yu2019deep, li2019relation}. In these tasks, effective multi-modal interaction and comprehensive understanding of both visual and linguistic features are critical in achieving great performance. Some previous works employ masked word prediction (MWP) to achieve this goal, where a proportion of words in a sentence are randomly masked, and the masked words are predicted under the condition of visual inputs \cite{kim2021vilt, ge2022bridging, zhu2020actbert}. Most MWP methods directly predict the value of the token. In our work, instead of predicting the single token, we reconstruct the holistic representation of text embedding and measure the global similarity, leading to the comprehensive understanding of the entire sentence.

Moreover, recently popular vision-language pre-training models \cite{radford2021learning, li2022grounded, yao2022detclip, yao2021filip} have demonstrated the remarkable ability of contrastive learning in cross-modal representation learning. Motivated by their success, we incorporate the contrastive loss in our image-to-text decoder to facilitate text reconstruction. The experimental results reflect that the two components are highly complementary and effectively enhance the semantic clues in visual features.

\section{Method}
The Dual Multi-Modal Interaction (DMMI) network adopts the encoder-decoder paradigm, which is illustrated in Fig.~\ref{fig:framework}. In the encoder part, the visual encoder and text encoder are utilized to extract visual and linguistic features, respectively. During this process, the Multi-scale Bi-direction attention (MBA) module is employed to perform cross-modal interaction. After feature extraction, the two modalities are delivered to the decoder part. In the text-to-image decoder, the text embedding is utilized to query the visual feature and generate the segmentation mask. While in the image-to-text decoder, we employ the Context Clue Recovery (CCR) module to reconstruct the erased information of target entity conditioned on the visual features. Meanwhile, the contrastive loss is utilized to promote the learning of CCR module. We elaborate each component of the DMMI network in detail in the following sections.

\subsection{Feature Encoder}
Given the text expression $T$, we randomly mask the entity-phrase via TextBlob Tool \cite{loria2018textblob} and generate its corresponding counterpart $T^{\prime}$. Then, we feed both $T$ and $T^{\prime}$ into the text encoder to generate the linguistic features $E =\left\{ e_{l}\right\}^{L}_{l=1}$ and $ E^{\prime } =\left\{ e^{\prime }_{l}\right\}^{L}_{l=1} \in \mathbb{R}^{C_t \times L}$, where $C_t$ and $L$ indicate the number of channels and the length of the sentence. For the input image $X$, we utilize the visual encoder to extract the multi-level visual features $V_n \in \mathbb{R}^{C_n \times H_n \times W_n}$. Here, $C_n$, $H_n$ and $W_n$ denote the number of channels, height and width, and $n$ indicates features in the $n$-th stage. During the feature extraction, MBA module is hierarchically applied to perform cross-modal feature interaction.

\vspace{-2mm}
\subsubsection{Hierarchical Structure}
As illustrated in Fig.~\ref{fig:framework} (a), the visual encoder is implemented as a hierarchical structure with four stages, which is conducted with the MBA module alternately. For the shallow layer feature $V_1$ extracted from the first stage of visual encoder, we deliver it to MBA module with linguistic feature $E_1$ and obtain $V^{\ast }_{1}$ and $E^{\ast }_{1}$. Then, $V^{\ast }_{1}$ is sent back to the visual encoder, based on which $V_2$ is extracted through the next stage. Meanwhile, $E^{\ast }_{1}$ is also noted as the $E_{2}$ that will be utilized in the next MBA module. Similarly, $V_2$ and $E_{2}$ are fed to MBA module again, and the generated $V^{\ast }_{2}$ will be delivered to the next part of visual encoder. By doing so, the visual and linguistic features are jointly refined, achieving cross-modal alignment in both text-to-image and image-to-text directions.

\vspace{-2mm}
\subsubsection{Multi-scale Bi-direction attention Module}
The MBA module jointly refines the visual feature $V$ and linguistic feature $E$ to achieve text-to-image and image-to-text alignment. To simplify the notation, here we drop the subscript of features from different stages. Inspired by the success of self-attention \cite{vaswani2017attention}, most recent works utilize the cross-attention operation to perform the cross-modal feature interaction. During this process, the visual feature $V$ is first flattened to $\mathbb{R}^{C \times N}$, where $N = W \times H$. Then, the feature interaction is formulated as:
\begin{equation}
V^{\ast }=\text{softmax} (\frac{(W_{q}V)^{T}(W_{k}E)}{\sqrt{\hat{C} } } )\left( W_{v}E\right)^{T}  
\label{eq:norcross}
\end{equation}
where $W_q$, $W_k$ and $W_v$ are three transform functions unifying the number of channels to $\hat C$. However, Eq.~\ref{eq:norcross} only establishes the relationship between a single pixel and a single word. In fact, beyond the single point representation, local visual regions and text sequences also store critical information for the comprehensive understanding of multi-modal features. Following this idea, we design two alignment strategies in MBA to capture the relationship between visual features and text sequences in different local regions.


\textbf{Text-to-Image Alignment.}
To fully leverage the structure information in various regions, we compute the affinities coefficients $W^{\alpha,r}_{att}$ between each token and different local regions $\Omega^{\alpha}_{r}$, in which $r$ indicates different spatial sizes and its value ranges from 1 to $R$. $\Omega^{\alpha}_{r}$ will slide across the whole spatial plane of the visual feature. Then, given region $\Omega^{\alpha}_{r}\left( i\right)$ centered at the position $i$, the $i$-th row weight $w^{\alpha,r}_{i}$ in attention matrix $W^{\alpha ,r}_{att} \in \mathbb{R}^{N \times L}$ is calculated as:
\begin{equation}
w^{\alpha,r}_{i}=\text{softmax} (\sum_{m\in \Omega^{\alpha}_{r} \left( i\right)  } \frac{(W^{\alpha}_{q}V^{m})^{T}(W^{\alpha}_{k}E)}{\sqrt{\hat C} } )
\label{eq:wvisual}
\end{equation}
where $w^{\alpha,r}_{i} \in \mathbb{R}^{1 \times L}$, $m$ enumerates all spatial positions in $\Omega^{\alpha}_{r}\left( i\right)$ and $V^{m} \in \mathbb{R}^{\hat C \times 1}$ denotes one specific feature vector in $\Omega^{\alpha}_{r}\left( i\right)$. Then, for all $\Omega^{\alpha}_{r}$, the final affinities coefficient is calculated as:
\begin{equation}
W^{\alpha }_{att}=\sum^{R}_{r=1} \lambda^{\alpha }_{r} W^{\alpha ,r}_{att}
\label{eq:wvisavg}
\end{equation}
where $\lambda^{\alpha }_{r}$ is a learnable parameter reflecting the importance of regions in different sizes. Finally, after the process of transform function $W^{\alpha}_{v}$, the linguistic information is incorporated into the visual feature:
\begin{equation}
V^{\ast }=W^{\alpha}_{att}(W^{\alpha}_{v}E)^{T}
\label{eq:fuse1}
\end{equation}

\textbf{Image-to-Text Alignment.} In human perception, to fully comprehend the language expression, we will associate the context information rather than understanding each word separately. Therefore, for each visual pixel, we also establish the connection with various text sequences  $\Omega^{\beta}_{r}$, where $r$ indicates different lengths of the sequence and $\Omega^{\beta}_{r}$ slides across the whole sentence. For text sequence $\Omega^{\beta}_{r} \left( i\right) $ starting at position $i$, we calculate the $i$-th row weight $w^{\beta,r}_{i}$ in affinity coefficients $W^{\beta,r}_{att} \in \mathbb{R}^{L \times N}$ as:
\begin{equation}
w^{\beta,r}_{i}=\text{softmax} (\sum_{m\in \Omega^{\beta}_{r} \left( i\right)  } \frac{(W^{\beta}_{q}E^{m})^{T} (W^{\beta}_{k}V)}{\sqrt{\hat C} } )
\label{eq:wlin}
\end{equation}
where $w^{\beta,r}_{i} \in \mathbb{R}^{1 \times N}$, $m$ enumerates all tokens in $\Omega^{\beta}_{r} \left( i\right) $ and $E^{m} \in \mathbb{R}^{\hat C \times 1}$ represents one specific feature vector in $\Omega^{\beta}_{r} \left( i\right) $. Then, similar to Eq.~\ref{eq:wvisavg}, we average the $W^{\beta,r}_{att}$ through a set of learnable parameters $\lambda_{r}^{\beta}$ to obtain the $W^{\beta}_{att}$. Afterwards, the visual information is involved to generate the refined text embedding as follows:
\begin{equation}
E^{\ast }=W^{\beta}_{att}(W^{\beta}_{v}V)^{T}
\label{eq:fuse2}
\end{equation}

\subsection{Text-to-Image Decoder}
The whole structure of text-to-image decoder is depicted in Fig.~\ref{fig:framework}(b). As advocated in \cite{ronneberger2015u, chen2017rethinking}, we implement skip-connections between the encoder and decoder to introduce the spatial information stored in the shallow layers. Specifically, the text-to-image decoder can be described as:
\begin{equation}
\begin{cases}Y_{4}=V_{4}^{\ast}&\\ Y_{n}=\psi \left( \phi \left( Y_{n+1},V_{n}^{\ast}\right)  ,E_{4}^{\ast}\right)  &n=3,2\end{cases} 
\label{eq:vdecoder}
\end{equation}
in which $\psi \left( \cdot \right)  $ indicates the Transformer decoder layer. $\phi \left( \cdot \right)$ consists of two 3$\times$3 convolutions followed by batch normalization and the ReLU function, in which features from the shallow parts of the encoder are aggregated with the decoder feature. Then, a series of convolution operations are applied on $Y_2$ to produce two class score maps $\hat{Y}_1$, which is considered as the final visual prediction of DMMI network. Finally, we calculate the binary cross-entropy loss for $\hat{Y}_1$ with $Y_{gt}$, which is denoted as $\mathcal{L}_{ce}$.

\subsection{Image-to-Text Decoder}
\subsubsection{Context Clue Recovery Module}


Besides the text-to-image decoder, DMMI network promotes the referring segmentation in a dual manner and facilitates the information flow from visual to text, which is illustrated in Fig.~\ref{fig:framework}(c). For the incomplete linguistic feature $E^{\prime } =\left\{ e^{\prime }_{l}\right\}^{L}_{l=1}$, we utilize CCR module to reconstruct its masked information under the guidance of visual feature $V_g^{\ast}$. To support the precise reconstruction, the visual feature is encouraged to contain essential semantic clues stored in the $E =\left\{ e_{l}\right\}^{L}_{l=1}$, which boosts the sufficient multi-modal interaction in the encoder part and support the accurate segmentation in the text-to-image decoder.

Specifically, given the visual feature $V_{g}^{\ast}$, we employ a Transformer decoder layer $\mathcal{D}(E^{\prime }, V_{g}^{\ast})$ to recover the missed information in the $E^{\prime } =\left\{ e^{\prime }_{l}\right\}^{L}_{l=1}$, where visual feature $V_g^{\ast}$ is employed to query the $E^{\prime}$. Notably, we extract $V_{g}^{\ast}$ from middle part of the text-to-image decoder, which contains both spatial and semantic information. The output of $\mathcal{D}(E^{\prime }, V_{g}^{\ast})$ is considered as the reconstructed text embedding, which is denoted as $\hat{E} =\left\{ \hat{e}_{l}\right\}^{L}_{l=1}$.

To enforce the CCR module to precisely recover the missing information, we measure the similarity distance between the reconstructed embedding $\hat{E}$ and $E_4^{\ast}$, and calculate $\mathcal{L}_{sim}$ as:
\begin{equation}
\small
\mathcal{L}_{\text {sim }}=\delta *\left(1-\cos \left[\operatorname{Detach}\left(E_4^*\right), \hat{E}\right]\right)
\label{eq:losssim}
\end{equation}
Here, $\delta$ is an indicator that will be set to 0 if this sample is a one-to-zero case, where the text input is unrelated to the corresponding image, making it impossible to reconstruct linguistic information. Additionally, $\operatorname{Detach}(E_{4}^{\ast})$ refers to stopping the gradient flow of $E_4^{\ast}$ in Eq.~\ref{eq:losssim}, which prevents $E_4^{\ast}$ from being misled by $\hat{E}$ in the optimization. 
\vspace{-3.5mm}
\subsubsection{Multi-modal Contrastive Learning}
\vspace{-1.5mm}
We calculate the contrastive loss to reduce the distance between visual feature and its corresponding linguistic one, which is helpful in reconstructing the text embedding from the visual representation. Specifically, we aggregate features from different parts of text-to-image decoder to generate $\tilde V_d^{\ast} $. Then, for visual feature $ \tilde V_d^{\ast} \in \mathbb{R}^{\mathcal{B} \times N \times C}$ and linguistic feature $ \tilde E_4^{\ast} \in \mathbb{R}^{\mathcal{B} \times L \times C}$ in a batch, we pool them into $V_o$ and $E_o\in \mathbb{R}^{\mathcal{B} \times C}$. Afterwards, the contrastive loss is computed as:

\vspace{-2mm}
\begin{equation}
\mathcal{L}_{con} = \mathcal{L}_{I \rightarrow T}+\mathcal{L}_{T \rightarrow I}\
\label{eq:losscon}
\end{equation}
where $\mathcal{L}_{I \rightarrow T}$ and $\mathcal{L}_{T \rightarrow I}$ denote image-to-text and text-to-image contrastive loss respectively:
\vspace{-2mm}
\begin{equation}
\footnotesize
\mathcal{L}_{I \rightarrow T}=-\frac{1}{\mathcal{B}} \sum_{i=1}^\mathcal{B} \delta^{(i)}*\log \frac{\exp \left(V_o^{(i)} \cdot E_o^{(i)}/ \tau\right)}{\sum_{j=1}^\mathcal{B} \exp \left(V_o^{(i)} \cdot E_o^{(j)} / \tau\right)}
\label{eq:losscon1}
\end{equation}
\begin{equation}
\footnotesize
\mathcal{L}_{T \rightarrow I}=-\frac{1}{\mathcal{B}} \sum_{i=1}^\mathcal{B} \delta^{(i)}*\log \frac{\exp \left(E_o^{(i)} \cdot V_o^{(i)}/ \tau\right)}{\sum_{j=1}^\mathcal{B} \exp \left(E_o^{(i)} \cdot V_o^{(j)} / \tau\right)}
\label{eq:losscon2}
\end{equation}
where $V_o^{(i)} \in \mathbb{R}^{C}$ and $E_o^{(i)} \in \mathbb{R}^{C}$ denote $i^{th}$ sample in a batch, $\mathcal{B}$ indicates the batch size. Meanwhile, $\delta$ is the one-to-zero indicator, $\tau $ is the temperature hyper-parameter that scales the logits. Finally, the total loss is combined as the summation of $\mathcal{L}_{ce}$, $\mathcal{L}_{sim}$ and $\mathcal{L}_{con}$ over the batch.

\vspace{-2mm}
\section{Ref-ZOM Dataset}
\vspace{-1mm}
We collect Ref-ZOM to address the limitations of mainstream datasets \cite{kazemzadeh2014referitgame, nagaraja2016modeling, mao2016generation} that only contain one-to-one samples. Following previous works \cite{kazemzadeh2014referitgame, nagaraja2016modeling, mao2016generation}, images in Ref-ZOM are selected from COCO dataset \cite{lin2014microsoft}. Generally, Ref-ZOM contains 55078 images and 74942 annotated objects, in which 43,749 images and 58356 objects are utilized in training, and 11329 images and 16,586 objects are employed in testing. Notably, Ref-ZOM is the first dataset that contains one-to-zero, one-to-one, and one-to-many samples simultaneously. It is worthwhile to mention that although the VGPHRASECUT dataset \cite{wu2020phrasecut} includes some one-to-many samples, it lacks one-to-zero cases, which makes it less applicable than Ref-ZOM. Due to the space limitation, we only illustrate a selection of representative samples from Ref-ZOM in Fig.~\ref{fig:refcocom}. More detailed information can be found in the supplementary materials.

\textbf{One-to-many.}
We collect one-to-many samples in three different ways, as illustrated in the first row of Fig.~\ref{fig:refcocom} from left to right. (1) We manually create some image-text pairs based on the expressions from COCO\_Caption and annotate the corresponding target in a two-player game \cite{kazemzadeh2014referitgame, yu2016modeling}. Specifically, given an image with caption expressions and annotations, the first annotator selects and modifies the sentence to describe the masked objects. Then, only given the image, the second annotator is asked to select the targets according to the text expression from the first one. The image-text pair will be collected only when the second annotator selects the targets correctly. (2) Based on the existing one-to-one referring image segmentation dataset, we combine the text expression describing different targets in one image to compose the one-to-many expressions. (3) We utilize the category information with the prompt template to compose some text samples. Generally, Ref-ZOM contains 41842 annotated objects under one-to-many settings.

\textbf{One-to-zero.} We carefully select 11937 images from COCO dataset \cite{lin2014microsoft}, which are not included in \cite{kazemzadeh2014referitgame, nagaraja2016modeling, mao2016generation}. Next, we randomly pair each image with a text expression taken from either the COCO captions or the text pools in \cite{kazemzadeh2014referitgame, nagaraja2016modeling, mao2016generation}. Finally, we conduct a thorough double-checking process to verify that the selected text expressions are unrelated to the corresponding images.

\textbf{One-to-one.} First, we randomly select some samples from existing datasets \cite{kazemzadeh2014referitgame, nagaraja2016modeling, mao2016generation}. Meanwhile, we manually  create some new samples based on the category information with the prompt template, which is similar with the third strategy in the creation of one-to-many samples. In total, there are 42421 one-to-one objects in the Ref-ZOM.

\begin{figure}[tbp] 
\centering 
\includegraphics[width=1\linewidth]{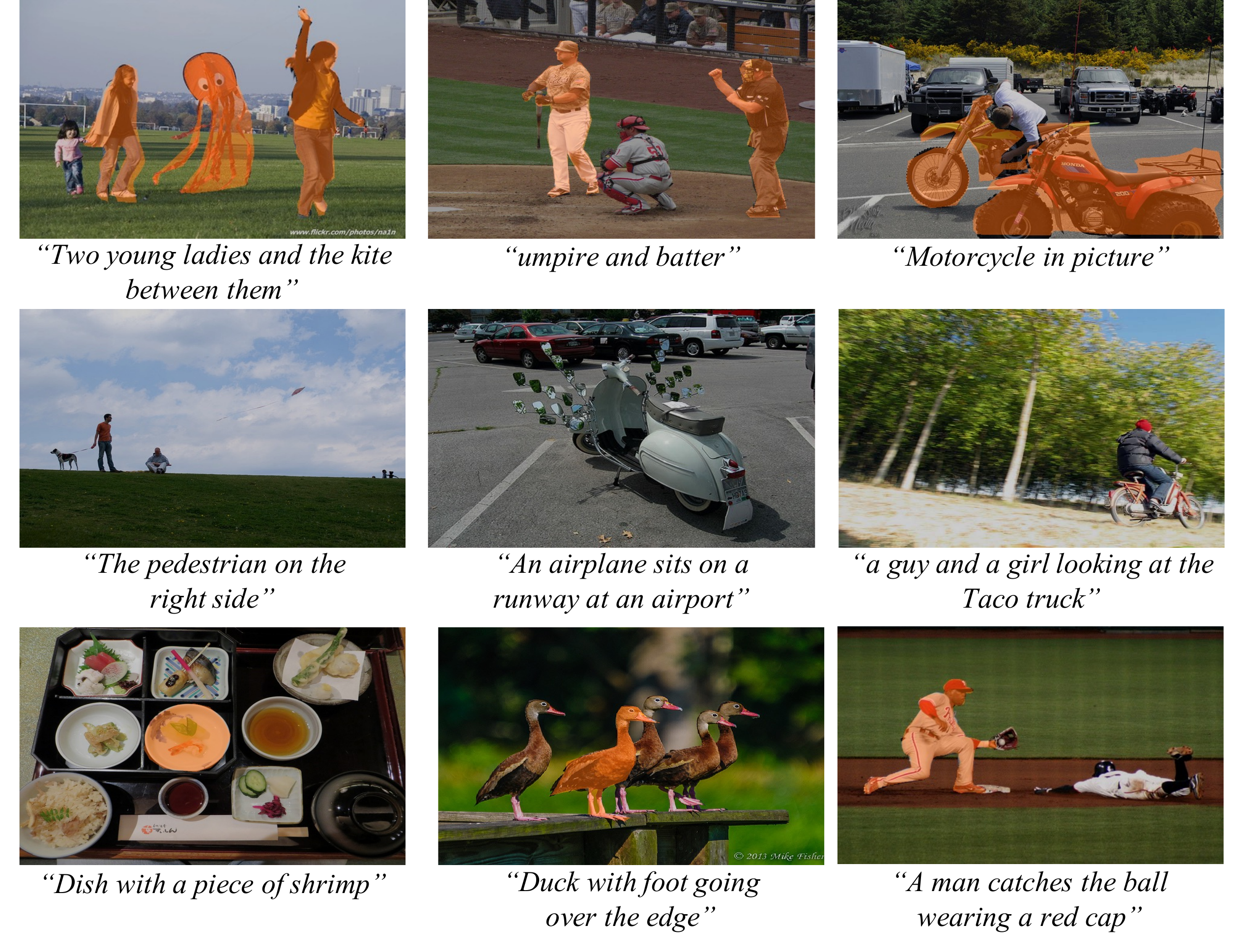}
\vspace{-7mm}
\caption{Selected samples from our newly collected Ref-ZOM datasets. From top to down are image-text pairs under one-to-many, one-to-zero and one-to-one settings.} 
\label{fig:refcocom} 
\vspace{-5mm}
\end{figure}

\vspace{-2mm}
\section{Experiments}
\subsection{Implementation Details}
We evaluate the performance of DMMI with two different visual encoders, ResNet-101 and Swin-Transformer-Base (Swin-B), which are initialized with classification weights pre-trained on ImageNet-1K and ImageNet-22K \cite{deng2009imagenet}, respectively. Our text encoder is the base\_BERT model with 12 layers and the hidden size of 768, which is initialized with the official pre-trained weights \cite{devlin2018bert}. In the training, we utilize AdamW as the optimizer with a weight decay of 0.01. Moreover, the initial learning rate is set to 5e-5 with a polynomial learning rate decay policy. The images are resized to 448 $\times$ 448 and the maximum sentence length is set to 20. Additionally, we only randomly erase one phrase in each iteration for each sentence.

In DMMI network, the values of $r$ for $\Omega^{\alpha}_{r}$ and $\Omega^{\beta}_{r}$ are set to [1, 2, 3]. Specifically, $\Omega^{\alpha}_{1}$, $\Omega^{\alpha}_{2}$, and $\Omega^{\alpha}_{3}$ correspond to spatial regions of size $1\times1$, $3\times3$, and $5\times5$, respectively. In addition, when $r=1, 2, 3$, $\Omega^{\beta}_{r}$ corresponds to text sequences with 1, 2, and 3 tokens, respectively. Furthermore, all Transformer layers in the decoder are set with 8 heads and temperature $\tau $ in $\mathcal{L}_{con}$ equals to 0.05.


\begin{table*}[tbp]
  \centering
  \small
  \caption{Comparison with state-of-the-art methods in terms of oIoU(\%) on three datasets. In G-Ref, U and G denote the UMD and Google partition, respectively. The best results are in bold.}
  \vspace{-2mm}
    \begin{tabular}{l|c|c|c|c|c|c|c|c|c|c}
    \toprule[1pt]
    \multirow{2}{*}{Method} & \multirow{2}{*}{\makecell[c]{Visual \\ Encoder}} & \multicolumn{3}{c|}{RefCOCO} & \multicolumn{3}{c|}{RefCOCO+} & \multicolumn{3}{c}{G-Ref} \\
\cline{3-11}      &    & val   & test A & test B & val   & test A & test B & val (U) & test (U) & val (G) \\
    \hline
    RRN \cite{huang2020referring} & ResNet-101 & 55.33  & 57.26  & 53.93  & 39.75  & 41.25  & 36.11  &  --   &  --  & 36.45  \\
    MAttNet \cite{yu2018mattnet} & ResNet-101 & 56.51  & 62.37  & 51.70  & 46.67  & 52.39  & 40.08  & 47.64  & 48.61  & -- \\
    CAC \cite{chen2019referring} & ResNet-101 & 58.90  & 61.77  & 53.81  &  --  &  --  &   -- & 46.37  & 46.95  & 44.32  \\
    LSCM \cite{hui2020linguistic} & ResNet-101 & 61.47  & 64.99  & 59.55  & 49.34  & 53.12  & 43.50  & --  &     -- & 48.05  \\
    CMPC+ \cite{liu2021cross} & ResNet-101 & 62.47  & 65.08  & 60.82  & 50.25  & 54.04  & 43.47  & -- &  --  & 49.89  \\
    MCN  \cite{luo2020multi} & DarkNet-53 & 62.44  & 64.20  & 59.71  & 50.62  & 54.99  & 44.69  & 49.22  & 49.40  & -- \\
    EFN \cite{feng2021encoder} & ResNet-101 & 62.76  & 65.69  & 59.67  & 51.50  & 55.24  & 43.01  & --  & --    & 51.93  \\
    BUSNet \cite{yang2021bottom} & ResNet-101 & 63.27  & 66.41  & 61.39  & 51.76  & 56.87  & 44.13  &  -- & --    & 50.56  \\
    CGAN \cite{luo2020cascade} & ResNet-101 & 64.86  & 68.04  & 62.07  & 51.03  & 55.51  & 44.06  & 51.01  & 51.69  & 46.54  \\
    LTS \cite{jing2021locate} & DarkNet-53 & 65.43  & 67.76  & 63.08  & 54.21  & 58.32  & 48.02  & 54.40  & 54.25  & -- \\
    VLT  \cite{ding2021vision} & DarkNet-53 & 65.65  & 68.29  & 62.73  & 55.50  & 59.20  & 49.36  & 52.99  & 56.65  & 49.76  \\
    \hline
    \rowcolor{mygray}
    \textbf{DMMI~(Ours)} & ResNet-101 & \textbf{68.56}  & \textbf{71.25}  & \textbf{63.16}  & \textbf{57.90}  & \textbf{62.31}  & \textbf{50.27}  & \textbf{59.02}  & \textbf{59.24}  & \textbf{55.13}  \\
    \hline
    \hline
    ReSTR \cite{kim2022restr} & ViT-B & 67.22  & 69.30  & 64.45  & 55.78  & 60.44  & 48.27  & --  &  -- & 54.48 \\
    CRIS \cite{wang2022cris} & CLIP-R101 & 70.47  & 73.18  & 66.10  & 62.27  & 68.08  & 53.68  & 59.87  & 60.36  & -- \\
    LAVT \cite{yang2022lavt} & Swin-B & 72.73  & 75.82  & 68.79  & 62.14  & 68.38  & 55.10  & 61.24  & 62.09  & 60.50  \\
    \hline
    \rowcolor{mygray}
    \textbf{DMMI~(Ours)} & Swin-B & \textbf{74.13}  & \textbf{77.13}  & \textbf{70.16}  & \textbf{63.98}  & \textbf{69.73}  & \textbf{57.03}  & \textbf{63.46}  & \textbf{64.19}  & \textbf{61.98}  \\
    \bottomrule[1pt]
    \end{tabular}%
  \label{tab:acccom1}%
\vspace{-4mm}
\end{table*}%

\begin{table}[tbp]
  \centering
  \small
  \caption{Comparisons with some representative methods on the newly collected Ref-ZOM dataset.}
  \vspace{-3mm}
    \begin{tabular}{l|c|c|c}
    \toprule[1pt]
    Method & oIoU  & mIoU  & Acc \\
    \hline
    MCN  \cite{luo2020multi} &  55.03 &  54.70  & 75.81 \\
    CMPC \cite{huang2020referring} &  56.19 &  55.72  & 77.01 \\
    VLT \cite{ding2021vision}  & 60.21 & 60.43 & 79.26 \\
    LAVT \cite{yang2022lavt}  & 64.45 & 64.78 & 83.11 \\
    \hline
    \textbf{DMMI~(Ours)}  & \textbf{68.77} & \textbf{68.21} & \textbf{87.02} \\
    \bottomrule[1pt]
    \end{tabular}%
  \label{tab:accrefm}%
  \vspace{-6mm}
\end{table}%

\vspace{-1mm}
\subsection{Datasets and Metrics}
\vspace{-1mm}
In addition to our newly collected Ref-ZOM, we evaluate our method on three mainstream referring image segmentation datasets, RefCOCO\cite{kazemzadeh2014referitgame}, RefCOCO+\cite{kazemzadeh2014referitgame} and G-Ref\cite{nagaraja2016modeling, mao2016generation}. Notably, G-Ref has two different partitions, which are established by UMD \cite{nagaraja2016modeling} and Google, respectively \cite{mao2016generation}. We evaluate our method on both of them.

In the test, for one-to-one and one-to-many samples, we employ the overall intersection-over-union (oIoU), the mean intersection-over-union (mIoU), and prec@$X$ to evaluate the quality of segmentation masks. The oIoU measures the ratio between the total intersection area and the total union area added from all test samples, while the mIoU averages the IoU score of each sample across the whole test set. Prec@$X$ measures the percentage of test images with an IoU score higher than the threshold $X\in \left\{ 0.5, 0.7, 0.9\right\}  $. As for the one-to-zero samples, since there is no target included in the image, IoU-based metrics are not applicable. Thus, we utilize image-level accuracy (Acc) to evaluate the performance. For each one-to-zero sample, its Acc value is 1 only when all points in the prediction mask are classified as the background. Otherwise, the Acc value is 0. We average the Acc value across the whole test set.

\vspace{-1mm}
\subsection{Comparison with State-of-the Arts}
In Table~\ref{tab:acccom1}, we compare the proposed DMMI network with the state-of-the-art methods on RefCOCO, RefCOCO+, and G-Ref in terms of the oIoU metric. The table is divided into two parts according to their visual encoder. The first part reports the performance of methods equipped with CNNs as the visual encoder, while the second part presents the methods using Transformer-based structure or pre-trained backbones beyond ImageNet as the visual encoder. Generally speaking, DMMI delivers the best performance in two conditions. Here, taking the second part as the example for analysis. On the RefCOCO dataset, we surpass the second-best method by 1.4\%, 1.31\%, and 1.37\% on val, testA, and testB subsets, respectively. On RefCOCO+ dataset, our DMMI network achieves a significant gain over the SOTA method, with increases of 1.84\%, 1.35\%, and 1.93\% on the val, testA, and testB subsets, respectively. On the UMD partition of G-Ref dataset, 2.22\% and 2.1\% oIoU improvements are obtained, while a 1.48\% increase is also observed on the Google partition. Such improvements are consistent in the first part of Table~\ref{tab:acccom1}.

In addition, we reproduce some representative methods on the newly collected Ref-ZOM dataset and evaluate our DMMI against these methods. The performance comparison is presented in Table~\ref{tab:accrefm}. Here, our DMMI is equipped with Swin-B as the visual encoder. As shown, our method achieves the best performance in handling the one-to-many and one-to-zero settings. To be more specific, DMMI outperforms the second-best method by 4.32\% and 3.43\% in terms of oIoU and mIoU. Moreover, in terms of the metric for one-to-zero samples, DMMI surpasses the secondary method by 3.91\% in Acc results.

\begin{figure*}[tbp] 
\centering 
\includegraphics[width=1\linewidth]{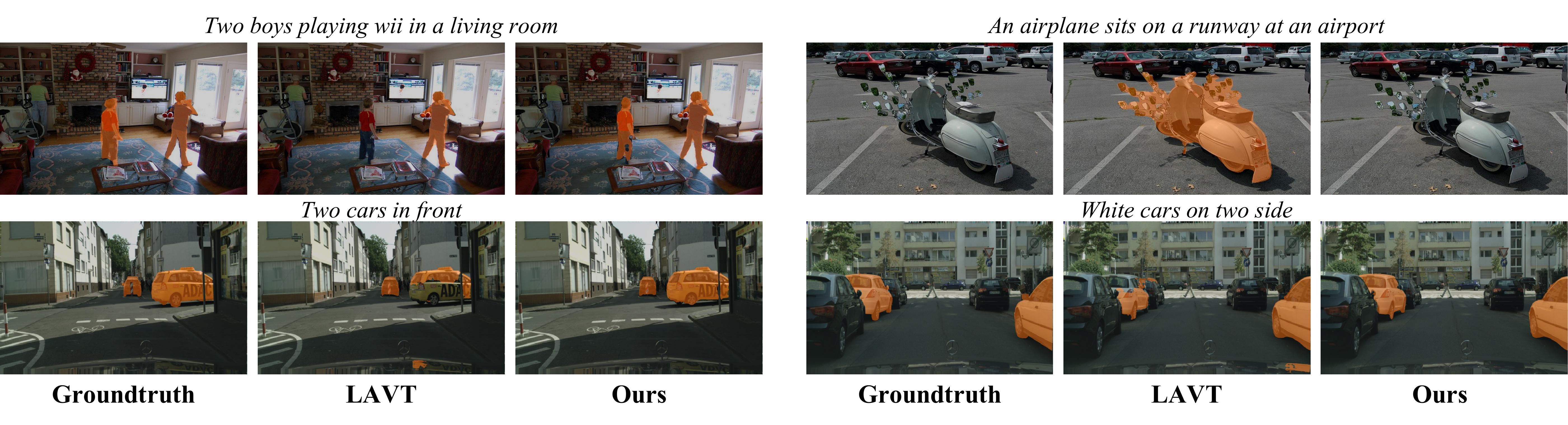}
\vspace{-8mm}
\caption{Comparisons of segmentation maps generated by LAVT and our DMMI network.} 
\label{fig:visual} 
\vspace{-4mm}
\end{figure*}

\vspace{-1mm}
\subsection{Ablation Study}
\vspace{-1mm}
In this part, we perform several ablation studies to evaluate the effectiveness of the key components in our DMMI network on both G-Ref\_(U) and Ref-ZOM datasets. The results are listed in Table~\ref{table:abla}. 

\textbf{Effect of Image-to-Text Decoder.} The first three rows in Table~\ref{table:abla} verify the effectiveness of the image-to-text decoder. First, we remove the whole image-to-text decoder and report the performance in the first row of Table~\ref{table:abla}, where a 1.7\% performance degradation could be observed on G-Ref. This reflects the image-to-text decoder contributes a lot to producing accurate segmentation result. Next, we add the similarity loss $\mathcal{L}_{sim}$ and report the results in the second row of Table~\ref{table:abla}. We can find $\mathcal{L}_{sim}$ brings significant improvements. Especially, on the Ref-ZOM, the accuracy improves by 1.48\% and 1.64\% in terms of oIoU and Acc. Meanwhile, 0.71\% oIoU gain is also found on G-Ref when the network is equipped with $\mathcal{L}_{sim}$. This demonstrates that through the reconstruction of incomplete text embedding in the training, DMMI is learned to fully incorporate the semantic clues about the entity targets into the visual features, which brings superior performance when facing various types of text inputs. Additionally, we verify the effectiveness of $\mathcal{L}_{con}$ in the third row. Compared to the baseline in the first row, performance goes up by 0.59\% and 0.82\% on the Ref-ZOM in terms of oIoU and Acc. This suggests $\mathcal{L}_{con}$ also contributes to the comprehensive understanding of target entity by pairing corresponding multi-modal features. Moreover, in the seventh row, we can find the best performance is achieved when the network equipped with $\mathcal{L}_{sim}$ and $\mathcal{L}_{con}$ simultaneously, reflecting the contrastive learning and text reconstruction are highly complementary.

\textbf{Effect of MBA module.} In the fourth to sixth row of Table~\ref{table:abla}, we conduct experiments to investigate the effectiveness of MBA module. On the one hand, as shown in the fifth and seventh row, if we prohibit the multi-modal interaction between various local regions, and only implement the interaction between single-word and single-pixel, the segmentation results drop significantly. Specifically, 1.46\% and 1.2\% degradation are observed on Ref-ZOM, which demonstrates the interaction in a large region benefits the comprehensive understanding of multi-modal features. On the other hand, we forbid the bi-direction mechanism in MBA module by removing the image-to-text alignment and only retaining the text-to-image one. The results are listed in the sixth row in Table~\ref{table:abla}, in which the performance drops a lot compared to the whole network. This reflects that mutually refining the multi-modal features in the interaction contributes to producing the accurate segmentation map.

\vspace{-1mm}
\subsection{Visualization}
\vspace{-1mm}
In this section, we visualize some segmentation maps generated from DMMI and LAVT \cite{yang2022lavt} to further demonstrate the superiority of our method. 

\begin{table}[tbp]
  \centering
  \small
  \caption{Ablation study of different components in DMMI network on G-Ref and Ref-ZOM datasets. Notably, ``Bi-D'' indicates the bi-direction operation in MBA module and I2T denotes the ``Image-to-Text''.}
  \vspace{-2mm}
    \begin{tabular}{r|cccc|c|cc}
    \toprule[1pt]
          & \multicolumn{2}{c|}{MBA} & \multicolumn{2}{c|}{I2T Decoder} & G-Ref & \multicolumn{2}{c}{Ref-ZOM} \\
\cline{2-8}          & Bi-D  & \multicolumn{1}{c|}{MS} & $\mathcal{L}_{sim}$   & $\mathcal{L}_{con}$   & oIoU  & oIoU  & Acc \\
    \hline
   1 &  \checkmark    &   \checkmark    &       &      &  61.76     &   65.77    & 83.91 \\
   2 &  \checkmark    &   \checkmark    &  \checkmark     &      &  62.47   &  67.25 & 85.55 \\
   3 &  \checkmark     &   \checkmark    &       &   \checkmark     &  62.13  &   66.36  & 84.73 \\
   \hline
   4 &      &     &   \checkmark    &    \checkmark   &  62.05     &  67.13  & 85.09 \\
   5 &  \checkmark   &       &   \checkmark    &    \checkmark   &  62.20     &  67.31  & 85.82 \\
   6 &    &    \checkmark   &   \checkmark    &    \checkmark   &  62.48     &   67.52  & 86.11 \\
    \hline
   7 &  \checkmark   &    \checkmark   &   \checkmark    &    \checkmark   &  \textbf{63.46}  &  \textbf{68.77}  &  \textbf{87.02} \\
    \bottomrule[1pt]
    \end{tabular}%
  \label{table:abla}%
  \vspace{-3mm}
\end{table}%


\textbf{Zero-shot to Ref-ZOM.} We first visualize some segmentation maps when the model is trained on the G-Ref and transferred to Ref-ZOM under the zero-shot condition. The results are illustrated in the first row of Fig.~\ref{fig:visual}. Since G-Ref only contains one-to-one samples, it is challenging to directly utilize the G-Ref-trained model to address the one-to-many and one-to-zero cases. As shown in the first sample, our DMMI network could precisely localize two boys and distinguish the women in the background. However, LAVT could only localize one boy with the largest size in the image. As for the second sample in the first row, DMMI also handles the one-to-zero case successfully.

\textbf{Zero-shot to Cityscapes.} To further verify the generalization ability of DMMI, we directly transfer the Ref-ZOM-trained networks to the Cityscapes dataset and give the model some expressions as the text input. The training images in Ref-ZOM all come from the COCO dataset, where the image style is quite different from that in Cityscapes. Thus, it is challenging to produce satisfactory performance when the model is transferred to Cityscapes without fine-tuning. As shown in the second row of Fig.~\ref{fig:visual}, DMMI presents the satisfactory performance. Specifically, when we give the text ``White cars on two side'', DMMI could precisely localize the corresponding targets while LAVT segment many irrelevant cars and fails to produce accurate segmentation map, demonstrating the great generalization ability of our method.

\vspace{-1mm}
\section{Conclusion}
\vspace{-1mm}
In this paper, we point out the limitations of existing referring image segmentation methods in handling expressions that refer to either no objects or multiple objects. To solve this problem, we propose a Dual Multi-Modal Interaction (DMMI) Network and establish the Ref-ZOM dataset. In the DMMI network, besides the visual prediction, we reconstruct the erased entity-phrase based on the visual features, which promotes the multi-modal interaction. Meanwhile, in the newly collected Ref-ZOM, we include image-text pairs under one-to-zero and one-to-many settings, making it more comprehensive than previous datasets. Experimental results show that the proposed method outperforms the existing method by a large margin, and Ref-ZOM dataset endows the network with remarkable generalization ability in understanding various text expressions. We hope our work provides a new perspective for future research.
\vspace{-1mm}
\section*{Acknowledgement} This paper is partially supported by the National Key R\&D Program of China No.2022ZD0161000 and the General Research Fund of Hong Kong No.17200622.


{\small
\bibliographystyle{ieee_fullname}
\bibliography{main}
}

\end{document}